%% file: root.tex
\documentclass[letterpaper, 10pt, conference]{ieeeconf}  

\IEEEoverridecommandlockouts                              

\usepackage{graphicx} 

\usepackage[
    colorinlistoftodos,
    prependcaption,
    textsize=tiny,
    obeyFinal,
]{todonotes}
\usepackage{subcaption}
\usepackage{tikz}
\usepackage{float}
\usepackage{amsmath}
\usepackage{xspace}

\usepackage{amssymb}

\usepackage{siunitx}
\usepackage{flafter}  
\usepackage[hidelinks]{hyperref}
\usepackage[
    nameinlink,  
    noabbrev,  
    capitalize,  
]{cleveref}  

\DeclareMathOperator*{\argmin}{arg\,min}
\DeclareMathOperator*{\argmax}{arg\,max}

\usepackage[sorting=none]{biblatex}
\usepackage{fancyhdr}
\fancypagestyle{withfooter}{
  
  \fancyfoot[C]{\footnotesize Accepted to the IEEE ICRA Workshop on Field Robotics 2024}
}

\title{\LARGE \bf
Situational Graphs for Robotic First Responders: \\ an 
application to dismantling drug labs
}

\author{W.J. Meijer\textsuperscript{\textsection}, A.C. Kemmeren\textsuperscript{\textsection}, J.M. van Bruggen, T. Haije, J.E. Fransman, J.D. van Mil
    \thanks{
    *This work is supported by TNO and Dutch National Police.
    }
    \thanks{\textsection These two authors contributed equally to this work.}
}

\begin{document}


\maketitle
\thispagestyle{withfooter}
\pagestyle{withfooter}

\input{src/abstract}

\input{src/config/symbols} 
\input{src/config/names} 

\input{src/introduction}

\input{src/related_work}
\input{src/problem}

\input{src/method}

\input{src/experiments}
\input{src/results}

\input{src/conclusion}
\printbibliography



\end{document}

%% file: src/abstract.tex
\begin{abstract}
In this work, we support experts in the safety domain with safer dismantling of drug labs, by deploying robots for the initial inspection.
Being able to act on the discovered environment is key to enabling this (semi-)autonomous inspection, e.g. to open doors or take a closer at suspicious items. Our approach addresses this with a novel environmental representation, the Behavior-Oriented Situational Graph, where we extend on the classical situational graph by merging a perception-driven backbone with prior actionable knowledge via a situational affordance schema.
Linking situations to robot behaviors facilitates both autonomous mission planning and situational understanding of the operator. 
Planning over the graph is easier and faster, since it directly incorporates actionable information, which is critical for online mission systems.
Moreover, the representation allows the human operator to seamlessly transition between different levels of autonomy of the robot, from remote control to behavior execution to full autonomous exploration.
We test the effectiveness of our approach in a real-world drug lab scenario at a Dutch police training facility using a mobile Spot robot and use the results to iterate on the system design.

\end{abstract}

%% file: src/config/symbols.tex
\newcommand{\mygraph}{\mathcal G}
\newcommand{\mygraphPost}{\mygraph_{post}}
\newcommand{\nodes}{\mathcal V}
\newcommand{\node}{v}
\newcommand{\edges}{\mathcal E}
\newcommand{\edge}{e}

\newcommand{\nodeSource}{\node_\text{source}}
\newcommand{\nodeTarget}{\node_\text{target}}
\newcommand{\weightMapping}{f}
\newcommand{\weight}{w}

\newcommand{\id}{\texttt {id}}
\newcommand{\position}{\boldsymbol{p}}

\newcommand{\agents}{\mathcal A}
\newcommand{\agent}{a}

\newcommand{\capabilities}{\mathcal K}
\newcommand{\behaviorCapabilities}{\capabilities_\behavior}
\newcommand{\agentCapabilities}{\capabilities_{a}}
\newcommand{\capabilitiesAgent}{\capabilities_\agent}
\newcommand{\capability}{k}

\newcommand{\behaviors}{\mathcal B}
\newcommand{\behavior}{b}
\newcommand{\behaviorDescriptor}{d^i_\behavior}

\newcommand{\cost}{C}

\newcommand{\objectives}{\mathcal O}
\newcommand{\objective}{o}
\newcommand{\reward}{R}
\newcommand{\rewards}{r}

\newcommand{\tasks}{\mathcal T}
\newcommand{\task}{\tau}
\newcommand{\taskOptimal}{\task^*}

\newcommand{\affordance}{h}
\newcommand{\affordances}{\mathcal H}

\newcommand{\situations}{\mathcal S}
\newcommand{\situation}{s}

\newcommand{\localmap}{M}

\newcommand{\objects}{\mathcal O}
\newcommand{\object}{o}

\newcommand{\objectlabel}{l}
\newcommand{\state}{x}
\newcommand{\statespace}{\mathcal{X}}

\newcommand{\plan}{\pi}
\newcommand{\plans}{\mathcal{P}}
\newcommand{\planOptimal}{\pi^*}

\newcommand{\taskAllocationProblem}{\mathbb T}

\newcommand{\graphSearchAlgo}{F}
\newcommand{\filter}{S}

%% file: src/config/names.tex
\newcommand{\sgraph}{Situational Graph\xspace}

%% file: src/introduction.tex

\section{Introduction}
\begin{figure}[t!]
    \begin{subfigure}{\columnwidth}
      \centering
      \includegraphics[width=0.95\columnwidth]{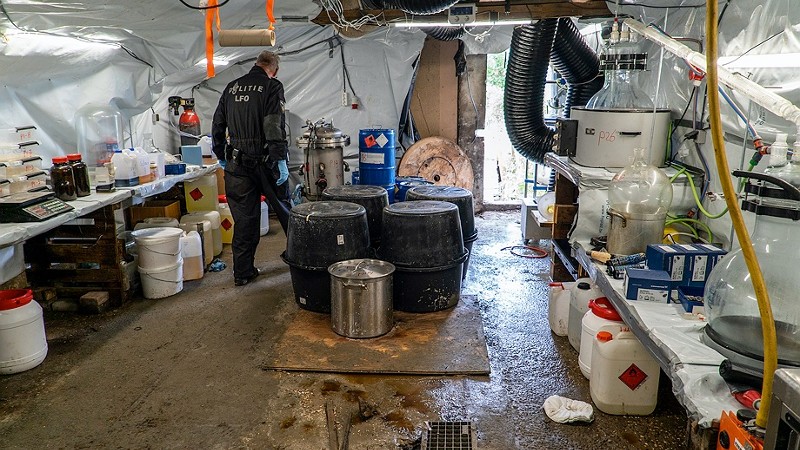}
    \end{subfigure} \\[3pt] %
    \begin{subfigure}{\columnwidth}
      \centering
      \includegraphics[width=0.95\columnwidth]{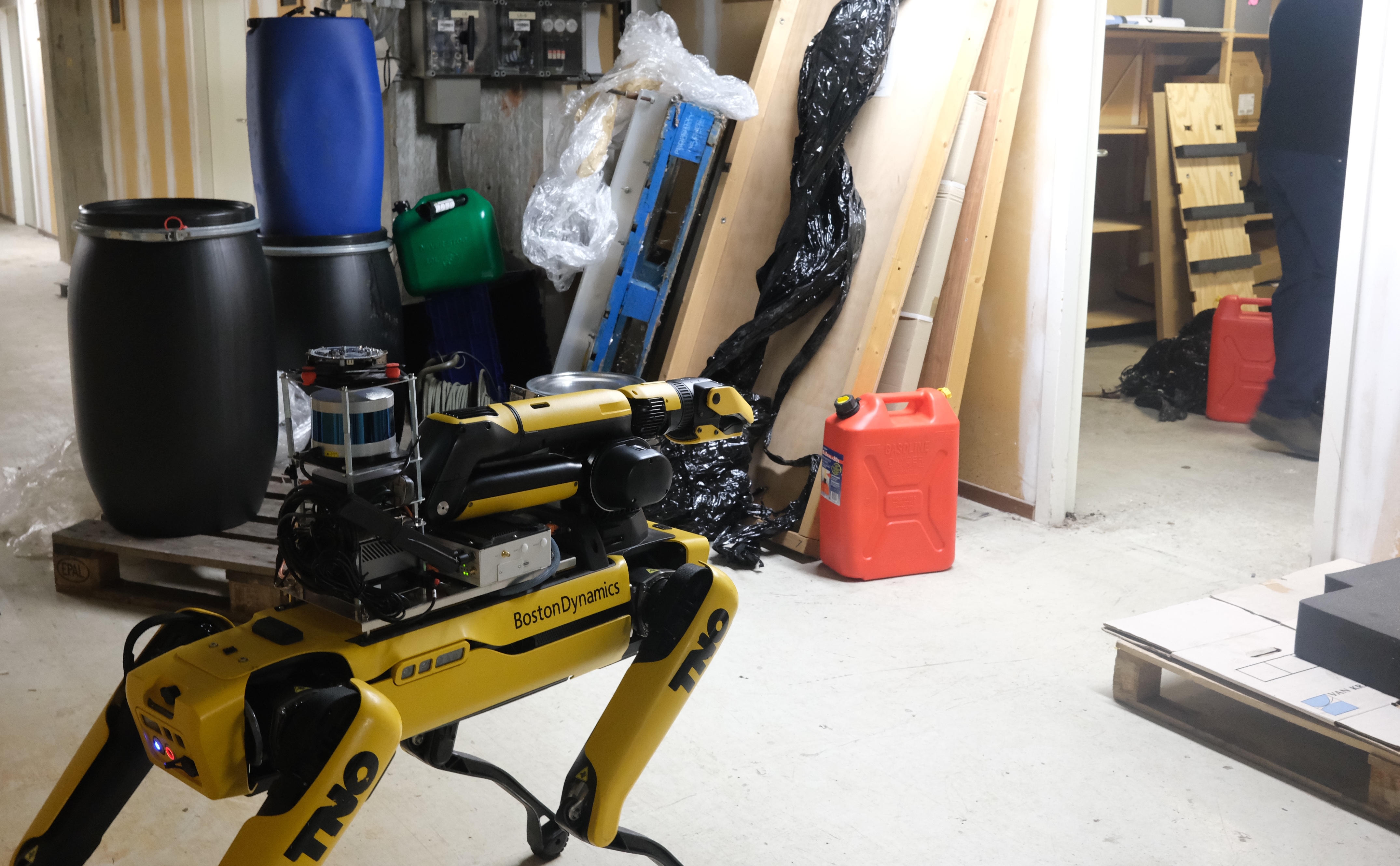}
    \end{subfigure}%
    \caption{Top: photo from a real drugslab (source:politie.nl). Bottom: Robot in operation during experiment in mock-up drugslab.
    }\vspace{-5mm}
    \label{fig:sfeer-impressie}
\end{figure}

\begin{figure}[t!]
    \centering
    \includegraphics[width=1\columnwidth]{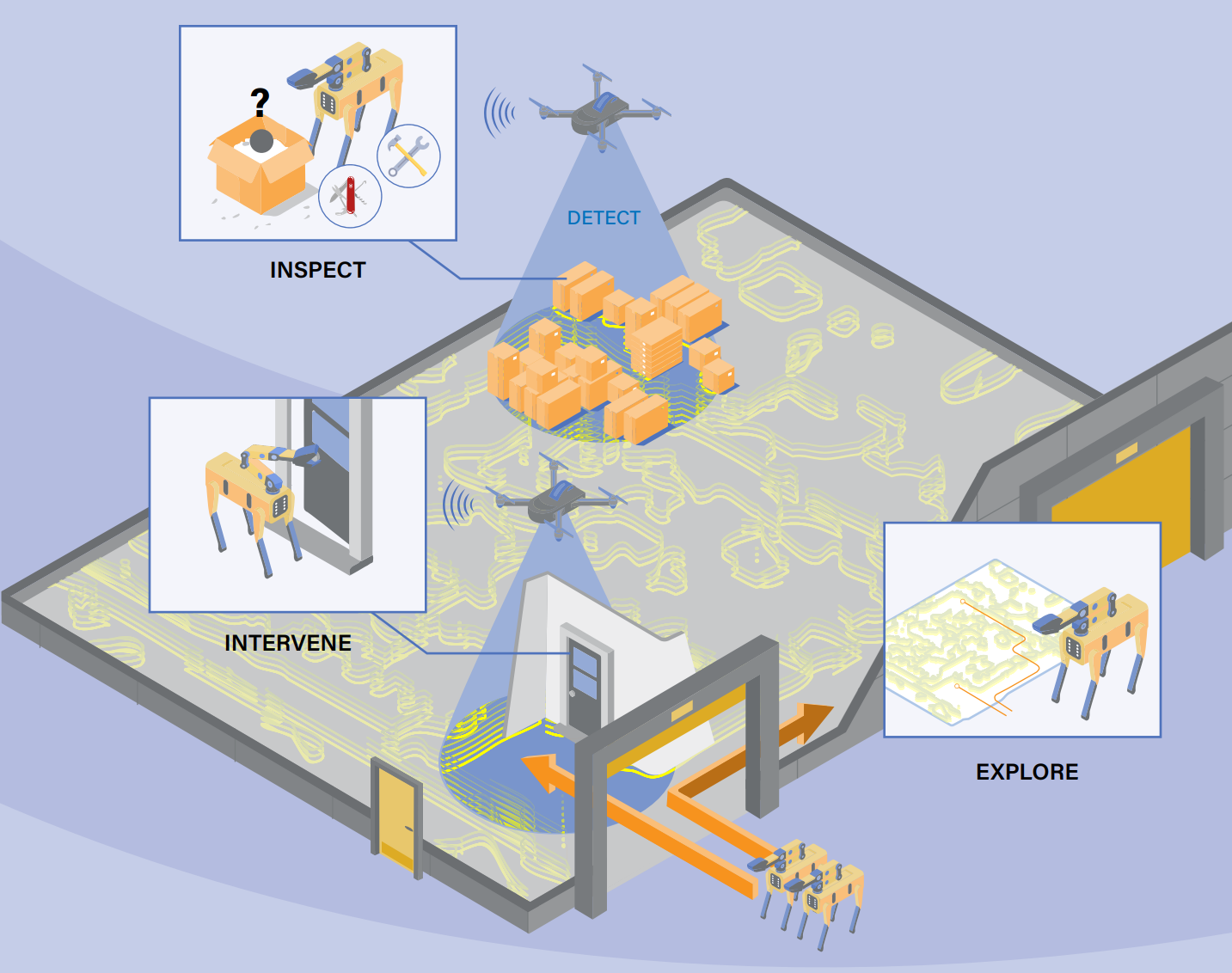}
         \caption{The envisioned usage of robots in a drug lab exploration scenario, including improvements identified in this formative study. Multiple robots collaborate to explore the environment, intervene on obstacles such as closed doors, and execute mission-relevant jobs such as inspection.   
     }
    \label{fig:concept-sketch}
\end{figure}

Illegal drug labs pose a difficult problem for the Dutch society and police force. 
The labs are increasingly found in urbanized areas, posing a high risk to local residents, and dumped drug waste has devastating effects on the surrounding nature\footnote{\url{https://open.overheid.nl/documenten/50b66102-48f4-4afe-a2ad-bbe034896e2f/file} (Dutch)}. 
Naturally, quick and safe dismantling of these labs is a priority. 
Various toxic or explosive chemicals and even suspects can be present in the labs, making dismantling a dangerous and complex mission. 
It is therefore important that the human specialists carefully plan these dismantling efforts since oversights have previously led to (severe) injuries of the staff. 
A potential solution to improve the safety of operation is to use a robotic system for the initial inspection of the drug lab (\cref{fig:sfeer-impressie}).
As illustrated in \cref{fig:concept-sketch}, robots could autonomously explore the lab and build a situational awareness - incl. detection of dangerous substances - which could then be used by the police staff to make a plan to properly dispose of any chemicals.

Unfortunately, commercially available robots are currently not yet suitable to support the police in this.
Companies providing state-of-the-art solutions focus on using autonomous robot platforms for routine inspection tasks: first, an operator records a mission in a known environment (e.g. a factory) through remote-control, which the robot can then replay\footnote{\href{https://bostondynamics.com/case-studies/energy-savings-predictive-maintenance-at-ab-inbevs-largest-european-brewery}{{\texttt{www.bostondynamics.com/case-studies/energy-savings\\-at-ab-inbevs-largest-european-brewery}}}}.

Inspecting a drug lab for safe dismantling does not fit this set-up, since the main aim is gaining situational awareness by the exploration of an unknown environment as opposed to performing the same known job instances routinely. 

The main challenge to deploy autonomous robotic systems for exploration of drug labs is to have a predictable and reliable system in an unknown environment. 
This requires a design where a human operator has insight into and control over the behavior of the robotic system at all times. 
We propose to use the Behavior-Oriented \sgraph to achieve these goals. 
The graph can be updated and extended in real-time upon exploration of new areas, while intuitively presenting any salient information in the environment such as tracked objects and traversability of the terrain. 
Moreover, it provides a representation that human operator and (autonomous) planner can simultaneously use to determine what behavior can be executed next. 
This allows for seamless handover between different levels of autonomy, from immersive remote-control to selecting specific autonomous behaviours (e.g. opening a door) to fully autonomous exploration.

%
This article shows our proposed framework, with the following contributions: 
\begin{enumerate}
    \item Initial design and formalization of a versatile environmental representation data structure that is directly actionable for robots and human operators, the Behavior-Oriented \sgraph;
    \item The process for real-time creation and adaptation of this \sgraph from data, which makes it suitable for missions in unknown environments;
    \item A workflow for operator interaction with this \sgraph during teleoperation to build situational awareness in high-stakes missions where the human should stay in control;
    \item Implementation of autonomous exploration and exploitation including the execution of behaviors, through algorithms that solve job selection and planning problems over the Behavior-Oriented \sgraph;
    \item A workflow where the shared control between human and autonomy is facilitated, by expressing missions in terms of the \sgraph.
\end{enumerate}
In collaboration with the Dutch police force, we are actively testing and iterating on this design. 
Using an initial version of the system, we gather feedback from potential users in a field test in a mock-up drug lab, resulting in iteration on requirements and suggested improvements.

%% file: src/related_work.tex
\section{Related Work}\label{sec:related-work}


The DARPA SubT challenge\footnote{\url{https://www.darpa.mil/program/darpa-subterranean-challenge}} spurred applied scientific research on building situational awareness in challenging unknown environments \cite{biggieFlexibleSupervisedAutonomy2023}.
The focus was to aid the situational awareness of an operator by reconstructing a 3D representation of the environment and localizing objects with decimeter accuracy in environments spanning kilometers.
However, where the DARPA SubT challenge solely considered `passive' exploration, new challenges arise when collecting information through active interaction with the environment. 
The following sections consider different methods to support active interactions, with methods that determine which actions can be executed where, and system designs that allow for adjustment of the level of autonomy. 

\subsection{Actionable Environment Representation}
Environment representations have been steadily improving in terms of information richness and the situational awareness they provide both for the robot itself and human operators \cite{bavleSLAMSituationalAwareness2023}.
The most prominent related field is Simultaneous Localization and Mapping (SLAM), which builds geometric representations of the environment \cite{shanLIOSAMTightlycoupledLidar2020}.
Current state-of-the-art representations extend regular SLAM pipelines to form their Situational Graphs \cite{bavleSituationalGraphsRobot2022}. These capture walls as conceptual geometric features and combine them to arrive at a conceptual understanding of rooms and floors which make up buildings.
Other methods such as 3D scene graphs (3DSG) \cite{rosinolKimeraSLAMSpatial2021} include relations between objects but don't explicitly model walls, rather relying on clustering of nodes to form rooms.  
However, the high dimensionality of these representations often make them unusable for planners. Even for simple tasks, the computational burden explodes with the size and complexity of the scene \cite{agiaTASKOGRAPHYEvaluatingRobot}.
%

Instead, other works focus on sparser graphs to encode actionable information, e.g. \textcite{greyTraversingEnvironmentsUsing2016} dynamically constructed a \emph{possibility graph}, where they solely focused on low level actions for traversing environments. 
Moreover, \textcite{chen2019BehavioralApproachVisual} connected the observed environment to higher level navigational behaviors such as "follow the corridor" and "take a right turn". \textcite{looBehaviourGraphsSemanticcontextual} extended this by proposing behavior graphs that map these distinct navigation behaviors in an environment. Here, they use \emph{navigational affordances} to refer to the system seeing which behaviors are relevant in the environment.

Along the same lines as these studies, we propose a robot-centric graph that encodes behaviors and ignores object-object relations to simplify the building and use of the graph. 
Similarly, we draw inspiration from 3DSG and add objects to the graph to make it more diversely applicable, and enable us to enrich the graph with detected affordances \cite{liOneShotOpenAffordance2023}. 

\subsection{Adjustable Level of Autonomy}\label{sec:adjustable-autonomy}
Depending on the complexity of the task, environment, and system, the operator may have varying levels of trust in the autonomous capabilities of the robot \cite{durstLevelsAutonomyAutonomous2014}.
This highlights the importance for an operator to control or aid the system during operation by interacting at varying levels. For example, \cite{adamsCanSingleHuman2023,galatiAuctionBasedTaskAllocation2023} use human supervision for a high-level task selection problem, where the human operator is able to edit the selected tasks. 
Additionally, \cite{selvaggioAutonomyPhysicalHumanRobot2021} supervises low-level motion control directly. 
When engaging in teleoperation of the system, it is beneficial to provide an immersive and intuitive experience for the operator
\cite{vanbruggenIBoticsAvatarSystem,vanerpWhatComesTelepresence2022}.
Transferring sensory modalities such as vision, audio or force feedback give the operator a feeling of being in control and contributes to the effectiveness of the operator \cite{walkerCyberPhysicalControlRoom2024}.
%
Moreover, studies have also investigated how to integrate these different levels of human supervision to make a system where the level of autonomy is adjustable \cite{katyalApproachesRoboticTeleoperation2014, marturiAdvancedRoboticManipulation2016}.
These works inspired us to create a traditional 2D view as the interface to interact with the system at a high level and adjust between levels of autonomy, and an immersive mixed reality view for intuitive teleoperation. 

%% file: src/problem.tex
\section{Problem Statement}\label{sec:problem}
We consider the exploration of a large unknown area. A human operator stands remotely, and is in charge of a robot. We address this challenge by dividing it into the following three sub-problems:
\begin{enumerate}
    \item[A.] The robot perceives the environment and records this information in a Behavior-Oriented \sgraph. 
    \item[B.] The graph stores actionable information (i.e. affordances), which the planning layer uses to autonomously execute behaviors. 
    \item[C.] This \sgraph is presented to the operator in a graphical user interface (GUI), providing information about the mission in real-time and allowing the operator to control the mission.  
\end{enumerate}
We elaborate on these three sub-problems in the following sections and from now on we refer to the Behavior-Oriented \sgraph, simply as \sgraph.

\subsection{\sgraph}\label{sec:problem-sgraph}
\input{resource/sgraph_diagram}
Let us first consider the problem of constructing a \sgraph from robot perception. 
We define the graph as a directed multigraph $\mygraph$, of which a schematic view is shown in \autoref{fig:sgraph-diagram}.
\begin{equation}
        \mygraph := ( \nodes, \edges ),
\end{equation}
where $\nodes$ is defined as the set of nodes and $\edges$ as the set of edges. 
Every node $\node \in \nodes$ represents a place and consists of a pose $\position_\node\in SE(2)$ in the ground plane and a situation $\situation_\node$. Here, $SE(2)$ is the 2-dimensional Special Euclidean group. 
\begin{equation}
    \begin{aligned}
        & \node :=  ( \position_\node, \situation_\node), \quad \node \in \nodes.
    \end{aligned} 
\end{equation}
For each node, situation $\situation_\node$ stores data specific to a location, which is relevant to determine possible behaviors. 
In our definition, the situation contains an $m\times n$ local gridmap $\localmap_\node\in\mathbb{R}^{m\times n}$ and a set of objects $\objects_\node$ that are currently present at place $\node$.
\begin{equation}
    \begin{aligned}
        \situation_\node &:=  (\localmap_\node, \objects_\node) \\
        \object &:=  (\id_\object, \objectlabel_\object, \position_\object), \quad \object\in\objects_\node,
    \end{aligned} 
\end{equation}
where object $\object$ has a unique identifier $\id_\object$, a label of the object type $\objectlabel_\object$ (e.g. \textit{door} or \textit{person}), and pose $\position_\object\in SE(3)$. 

A directed edge $\edge_{ij}\in\edges$ connects two nodes $\node_i$ and $\node_j$ iff some behavior $\behavior\in\behaviors$ allows the robot to transition from $\node_i$ to $\node_j$.
In other words, $\node_i$ implicitly encodes behavior pre-conditions, and $\node_j$ the post-conditions. 
Multiple edges are allowed between two nodes $\node_i$ to $\node_j$, since there can be multiple behaviors $\behavior_k$ that have the same pre- and post-conditions. 
$\objects^k_i$ is the set of objects at $\node_i$ that are relevant for the execution of $\behavior_k$.
$\edge^k_{ij}$ is therefore defined as
\begin{equation}
    \begin{aligned}
        & \edge^k_{ij} :=  ( \behavior_k(\objects^k_i), \node_i, \node_j ), \quad \edge^k_{ij} \in \edges,\quad \node_i, \node_j \in \nodes.
    \end{aligned} 
\end{equation}




\begin{figure*}
    \centerline{\includegraphics[width=0.7\textwidth]{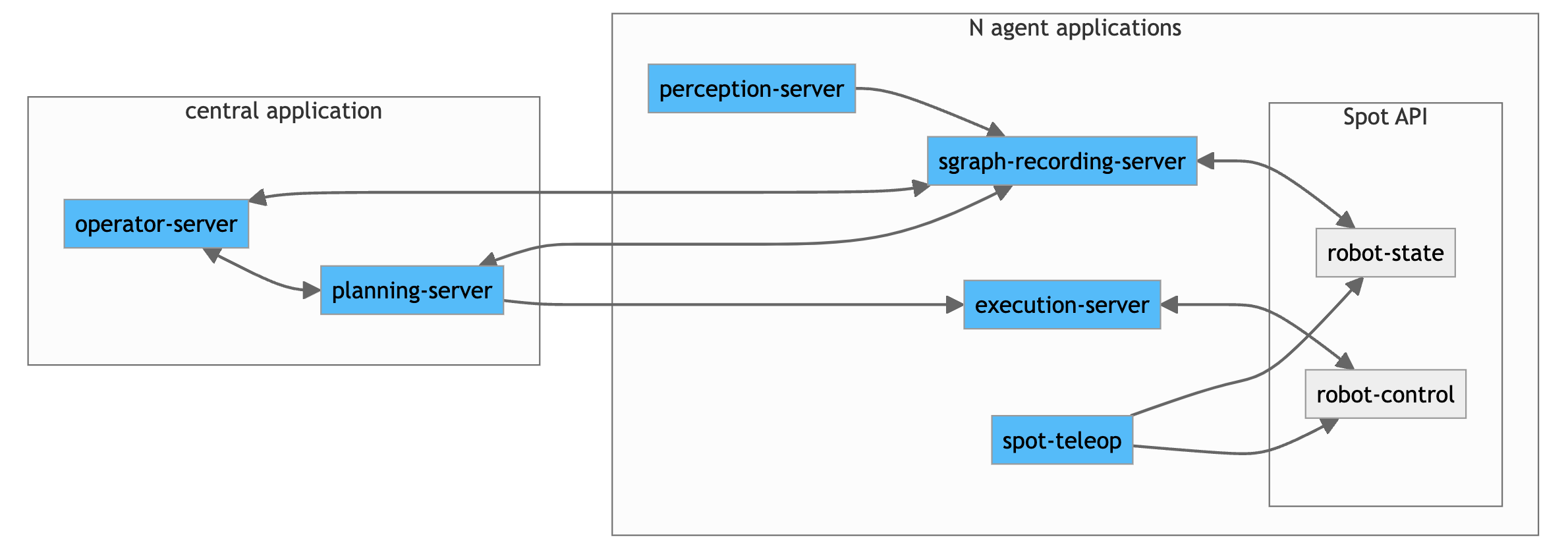}}
    \caption{
    A simplified overview of the centralized architecture of the system. 
    In the agent app running on the backpack PC a perceptions server publishes perception events from sensor data,
    which the sgraph-recording-server uses to update the \sgraph. 
    On a central app outside of the drug lab, the operator-server hosts the graphical user interface for the operator and allows the operator to select the level of autonomy.
    Based on this level, the planning-server and execution-server allocate and execute jobs respectively.
    Finally, when in teleop mode, the spot-teleop server allows direct communication with the spot API to improve teleop latency.
    }
    \label{fig:system-architecture}
\end{figure*}

We use the notion of affordance $\affordance$ to find the behavior that corresponds to an edge, which we define as a mapping 
\begin{equation}
        \affordance : \node_i \rightarrow \behavior_k, \node_j.
\end{equation}

Examples include detecting where new area's can be explored, or detecting that the environment can be changed by opening a door as shown in the top right of \cref{fig:sgraph-diagram}. 

Thus, based on the local gridmap $\localmap_{\node_i}$ and the objects $\objects_{\node_i}$ present in the local environment of $\node_i$, the robot infers which behaviors $\behavior_k$ 
are possible, and to what resulting desirable situation $\node_j$ this will lead. 
This way, we effectively encode one-step look-ahead planning into the graph.


\subsection{Planning Layer}\label{sec:method-planning}

The path planning problem concerns how some robot can reach node $\node_j$ from starting position $\node_i$. 
If the robot starts from node $\node_i$, the optimal plan $\planOptimal_{ij}$ to reach node $\node_j$ is simply the shortest path between $\node_i$ and $\node_j$ in the graph.
\begin{equation}
    \begin{aligned}
        & \planOptimal_{ij} :=  \argmin_{\plan_{ij}} \sum_{\edge\in\plan_{ij}} \weight_\edge, 
        \quad \plan_{ij}\in\plans_{ij} \\
    \end{aligned}
    \label{eq:optimal-plan}
\end{equation}
where $\weight_\edge$ is the cost of executing the behavior corresponding to edge $\edge$, and $\plans_{ij}$ is the set of all (acyclic) paths between nodes $\node_i$ and $\node_j$ in graph $\mygraph$. 
A plan is thus an ordered sequence (i.e. a path) of directed edges, encoding consecutive behaviors for the robot to traverse the graph. 

The job selection problem concerns selecting the optimal job $\taskOptimal$ given the agent's current node $\node_i$.
We define the optimal job to be selected as
\begin{equation}
    \begin{aligned}
        \taskOptimal = &\argmax_{\task} \reward(\task) - \cost(\task, \node_i) 
    \end{aligned}
    \label{eq:optimal-job}
\end{equation}
where $\reward(\cdot)$ is the reward function and $\cost(\cdot)$ is the cost function for the robot at $\node_i$ to reach the selected node $\task$. 
The cost $\cost$ to execute the job is the same as the cost of the optimal plan $\planOptimal_{ij}$, given $\node_i$ is the current place of the robot, and $\node_j=\task$ the allocated node.

The proposed system then allows for easy switching between the following levels of autonomy:
\begin{enumerate}
    \item Full autonomy with job selection and path planning
    \item Job selection by human operator, but autonomous path planning
    \item Human operator directly selects a behavior to execute that is available at the current node
    \item Motion directly remotely controlled by the operator
\end{enumerate}

\subsection{Human Operator}
 The situational graph is presented to the operator in a GUI, with which we aim to achieve three goals: (1) facilitate the situational awareness of the operator, (2) provide the operator with insight into the expected mission progress via visualization of the job assignment, plans for job execution and behavior execution, and (3) presenting an intuitive interface to adjust the level of autonomy of the robot, switching from e.g. immersive tele-operation, to execution of a specific behavior or to full autonomous exploration.

These goals have been identified based on conversations with stakeholders, and experience with similar projects in TNO's department of human machine teaming. 

%% file: resource/sgraph_diagram.tex
\tikzstyle{wp} = [circle, minimum width=1.5cm, minimum height=1.5cm, text centered, draw=black, fill=red!30, font=\footnotesize]
\tikzstyle{wo} = [circle, minimum width=1.5cm, minimum height=1.5cm, text centered, draw=black, fill=blue!50, font=\footnotesize]
\tikzstyle{ft} = [circle, minimum width=1.5cm, minimum height=1.5cm, text centered, draw=black, fill=green!30, font=\footnotesize]
\tikzstyle{agent} = [circle, minimum width=1.5cm, minimum height=1.5cm, text centered, draw=black, fill=yellow!50, font=\footnotesize]

\tikzstyle{goto} = [<->, >=stealth, color=red, line width=5pt]
\tikzstyle{behavior} = [->, >=stealth, color=blue, line width=5pt]
\tikzstyle{explore} = [->, >=stealth, color=red, line width=5pt]


\tikzstyle{startstop} = [rectangle, rounded corners, minimum width=2cm, minimum height=1cm,text centered, draw=black, fill=white!30, font=\small]
\tikzstyle{io} = [trapezium, trapezium left angle=70, trapezium right angle=110, minimum width=3cm, minimum height=1cm, text centered, draw=black, fill=blue!30]
\tikzstyle{process} = [rectangle, minimum width=2cm, minimum height=1cm, text centered, draw=black, fill=orange!30, font=\small]

\newcommand{\gotoBehavior}{\behavior_\text{goTo}}

\begin{figure}[t]
    \centering
    \resizebox{0.40\textwidth}{!}{
    \begin{tikzpicture}[node distance=4cm, align=center]
        \node (wp1) [wp] {\Large $\node_1$};
        \node (wp2) [wp, above of=wp1] {\Large $\node_2$};
        \node (wp3) [wp, right of=wp2] {\Large $\node_3$};
        \node (wp4) [wp, above of=wp3] {\Large $\node_4$};
        \node (ft1) [wo, left of=wp2] {\Large $\object_1$ \\ \\ \Large (frontier)};
        \node (wo1) [wo, below right of=wp3] {\Large $\object_2$\\ \\\Large (chemical\\ \Large container)};
        \node (wo2) [wo, right of=wp4] {\Large $\object_3$\\ \\ \Large (door)};
        
        \draw [goto] (wp1) -- (wp2) node[midway, left] {\Large $\gotoBehavior$};
        \draw [goto] (wp2) -- (wp3) node[midway, above] {\Large $\gotoBehavior$};
        \draw [goto] (wp3) -- (wp4) node[midway, left] {\Large $\gotoBehavior$};
        \draw [behavior] (wp3) -- (wo1) node[midway, above right] {\Large $\behavior_\text{requestTeleop}$};
        \draw [behavior] (wp4) -- (wo2) node[midway, above] {\Large $\behavior_\text{openDoor}$};
        \draw [behavior] (wp2) -- (ft1) node[midway, above] {\Large $\gotoBehavior$};
        \node (agent1) [agent, left of=wp2, xshift=3.2cm, yshift=-0.8cm] {\large robot};

        \draw [->] (-4,-0.5) -- ++(0.5,0) node[right]{\footnotesize\textit{x}};
        \draw [->] (-4,-0.5) -- ++(0,0.5) node[above]{\footnotesize\textit{y}};
    \end{tikzpicture}
    }\Large 
    \caption{
        An example of how the Behavior-Oriented \sgraph $\mygraph$ encodes potential behaviors $\behavior_k$ in the environment.
        Conventional edges $\edge$ between two nodes are visualized in red.
        Behaviors that take world objects as parameters are visualised by the blue arrows, connecting a node $\node$ with a world object $\object$.
        The robot can e.g. goTo and explore a frontier ($\object_1$) at $\node_2$ , open a door ($\object_3$) at node $\node_4$ or request the operator for assistance on a chemical container ($\object_2$) from node $\node_3$.
    }
    \label{fig:sgraph-diagram}
\end{figure}

%% file: src/method.tex
\begin{figure*}[t]
    \centerline{\includegraphics[width=0.9\textwidth]{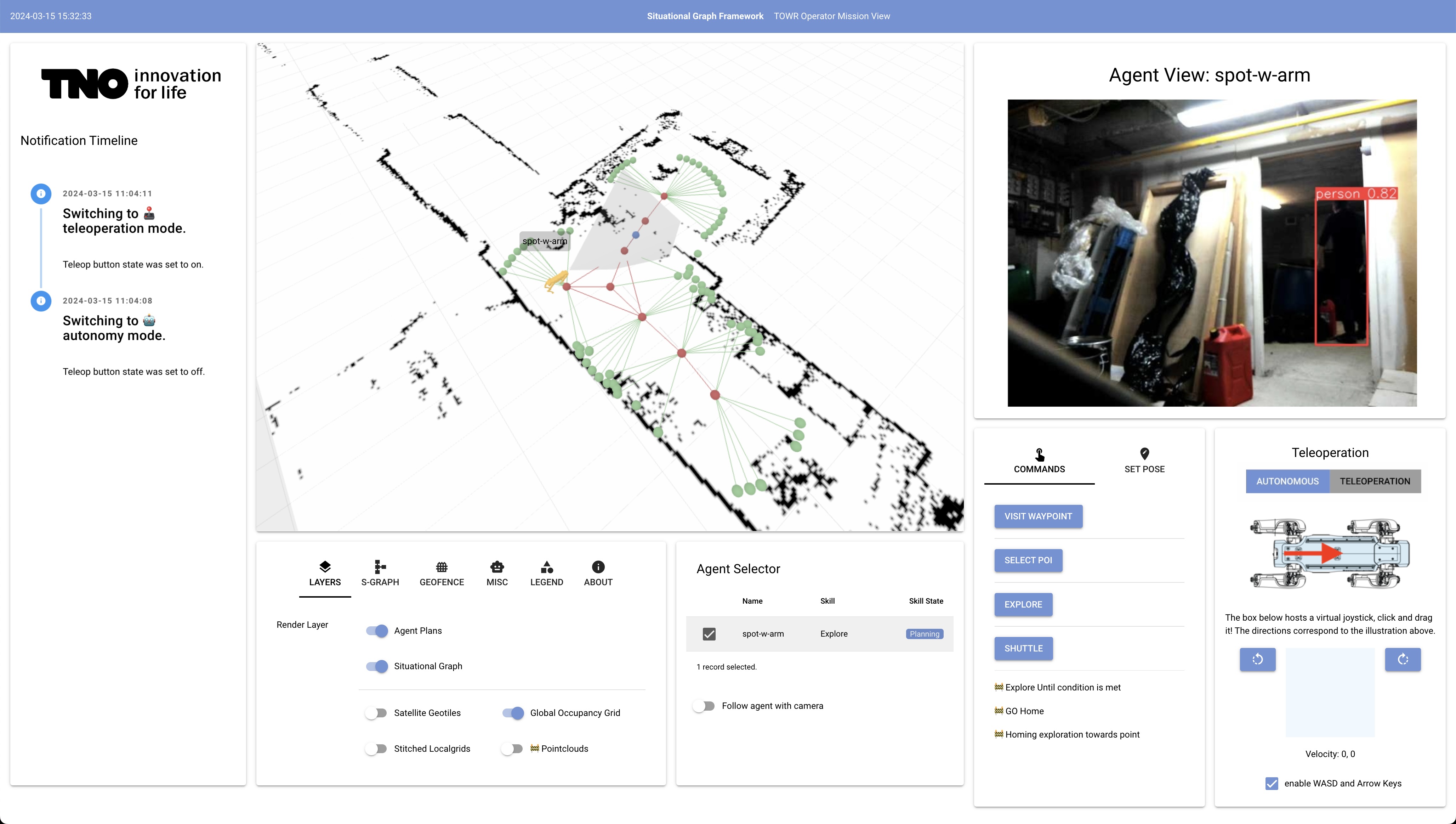}}
    \caption{The operator interface. The 3D scene in the center shows the \sgraph. Red nodes are waypoints visited by the robot, green nodes are unexplored frontiers and blue nodes are detected objects. Detections are represented by a unit vector from the position where the detection was done towards the detected object. The left side shows a timeline of events that occurred during the mission. The bottom elements house general options for the system, such as toggling layers on and off. The right side shows robot-specific elements such as the live stream (with detections), commands, and basic teleoperation controls.
    }
    \label{fig:gui}
\end{figure*}

\section{Methods and Algorithms}\label{sec:method}
In this section, an initial implementation of the \sgraph framework is presented.
To do this we first describe the servers for perception, \sgraph recording, planning, behavior execution, operator interaction, and immersive teleoperation.
The overall system architecture is outlined in \Cref{fig:system-architecture}.
The operator and planning server run centrally, while the rest run on the PC in the robot's backpack.

\subsection{Perception Server}\label{sec:methods-perception}
The perception server processes camera, IMU and LiDAR data in real-time to detect objects $\object$, update the node locations $\position_\node$ and produce gridmaps $\localmap$.

From video data, fiducials can be detected with highly accurate pose estimates and unique IDs. 
Moreover, a YOLOv8 model \cite{jocherUltralyticsYOLO2023} is able to detect a range of objects as defined in the COCO dataset \cite{linMicrosoftCOCOCommon2015}. YOLOv8's tracking capability is used to update object locations $\position_\object$ in real-time. 
As of yet, positions of the YOLOv8 detections are limited to image coordinates, and do not yet have 3D pose estimates. 

In parallel, data from a LiDAR and IMU are processed by the SLAM algorithm LIO-SAM \cite{shanLIOSAMTightlycoupledLidar2020}. 
This algorithm updates the robot pose used to update the node locations $\position_\node$ and processes 3D pointclouds to maintain a voxel map of the environment.
These pointclouds are also converted to local 2D gridmaps $\localmap_\node$ by using Octomap \cite{wurmOctomapProbabilisticFlexible}. 
A constructed global 2D map is shown in \autoref{fig:drugs-lab}. 

\subsection{\sgraph Recording}
The \sgraph Recording server has an \textit{observer module} and \textit{prediction module} to update the graph nodes and edges. 
The observer module receives perception events from the Perception Server.
The module uses these events to edit the graph by storing the latest gridmaps and object detections in the nodes and re-evaluating the validity of the edges. 

Furthermore, the robot uses the most recent estimate of its position to determine whether it is in a new place $\node_{new}\notin\nodes$. Here, a place is considered new if it is located $>\qty{2}{\meter}$ from the nearest node. Then, a new node is added to the graph and connected by an edge $\edge^k_{ij}$, where behavior $\behavior_k$ describes how the robot transitioned from $\node_i$ to the new node $\node_j$.

Then, the prediction module uses the recorded situation data and pre-defined affordances to further refine the graph, by detecting if a robot can execute any behaviors. However, this is limited to the behaviors and affordances defined before the mission. In our implementation, we can execute the following behaviors: 
\begin{description}
    \item[$\behavior_1$:] Go to node
    \item[$\behavior_2$:] Open a door
    \item[$\behavior_3$:] Request tele-operation from the human operator
\end{description}
Moreover, we have defined the following affordances $\affordance$ to map situations at node $\node_i$ to potential behaviors $\behavior_k$:
\begin{description}
    \item[$\affordance_1$:] If terrain in $\localmap_\node$ is traversable and leads to unknown areas, execute $\behavior_1$ to the frontier;
    \item[$\affordance_2$:] If a closed door is detected at some node, execute $\behavior_2$; 
    \item[$\affordance_3$:] If a human is detected, execute $\behavior_3$;
    \item[$\affordance_4$:] If a container is detected, execute $\behavior_3$.
\end{description}

Affordance $\affordance_1$ represents an important capability of our module, based on a frontier exploration algorithm. 
It checks whether any of the local grids afford traversable ground to go to unexplored areas. 
In those locations, new frontier nodes $\nodes_f\subset\nodes$ are sampled, while existing frontiers in recently explored areas are pruned.

\subsection{Planning Server}
The Planning Server considers two optimization problems: finding the best plan to execute a job (\cref{eq:optimal-plan}), and optimizing selection of the optimal job itself (\cref{eq:optimal-job}).

The path planning problem in \cref{eq:optimal-plan} is solved using Dijkstra's algorithm in NetworkX\footnote{\url{https://networkx.org/documentation/stable/reference/algorithms/shortest_paths.html}}. Furthermore, we define some positive reward $\reward$ for any frontier nodes $\nodes_f\subset\nodes$ such that the job selection problem in \cref{eq:optimal-job} reflects an exploration problem. 

It then depends on the selected level of autonomy (\autoref{sec:method-planning}) which of these problems are actively being solved. 
Both problems are optimized when the robot is in the highest level of autonomy, where it explores the environment without any human input. 
One level lower, the human operator selects the jobs, and the planning server only needs to consider the path planning problem. 
For the lowest two levels of autonomy, the planning server is inactive.


\subsection{Behavior Execution Server}
In the highest three levels of autonomy (\autoref{sec:method-planning}), the Behavior Execution Server can be called to execute the following behaviors: go to a node, open a door, or request teleoperation from a human operator. 

For the goTo node behavior a navigation stack is used, i.e. the Boston Dynamics navigation stack\footnote{\url{https://dev.bostondynamics.com/docs/concepts/autonomy/autonomous_navigation_services}} for Spot.
The behavior to open a door is currently only implemented for Spot, using a combination of the native Boston Dynamics API\footnote{\url{https://dev.bostondynamics.com/docs/concepts/arm/arm_services.html##door-service}} and custom detection of door handles and hinges. Lastly, the tele-operation request simply hands the control of the robot over to the operator.

\subsection{Operator Interface}
The operator interacts with the system using a web interface in the browser as shown in \autoref{fig:gui}. 
The view is split into generic elements on the left and robot-specific elements on the right. 
Various layers can be toggled on and off in this 3D view, for example specific node types of the \sgraph, a global 2D occupancy grid and plan visualizations. 
Also, the visualized \sgraph allows the human to interact with the system on a high level, since e.g. a job can be re-allocated by clicking on one of the nodes in the graph.

\begin{figure}
    \centering
    \includegraphics[width=\columnwidth]{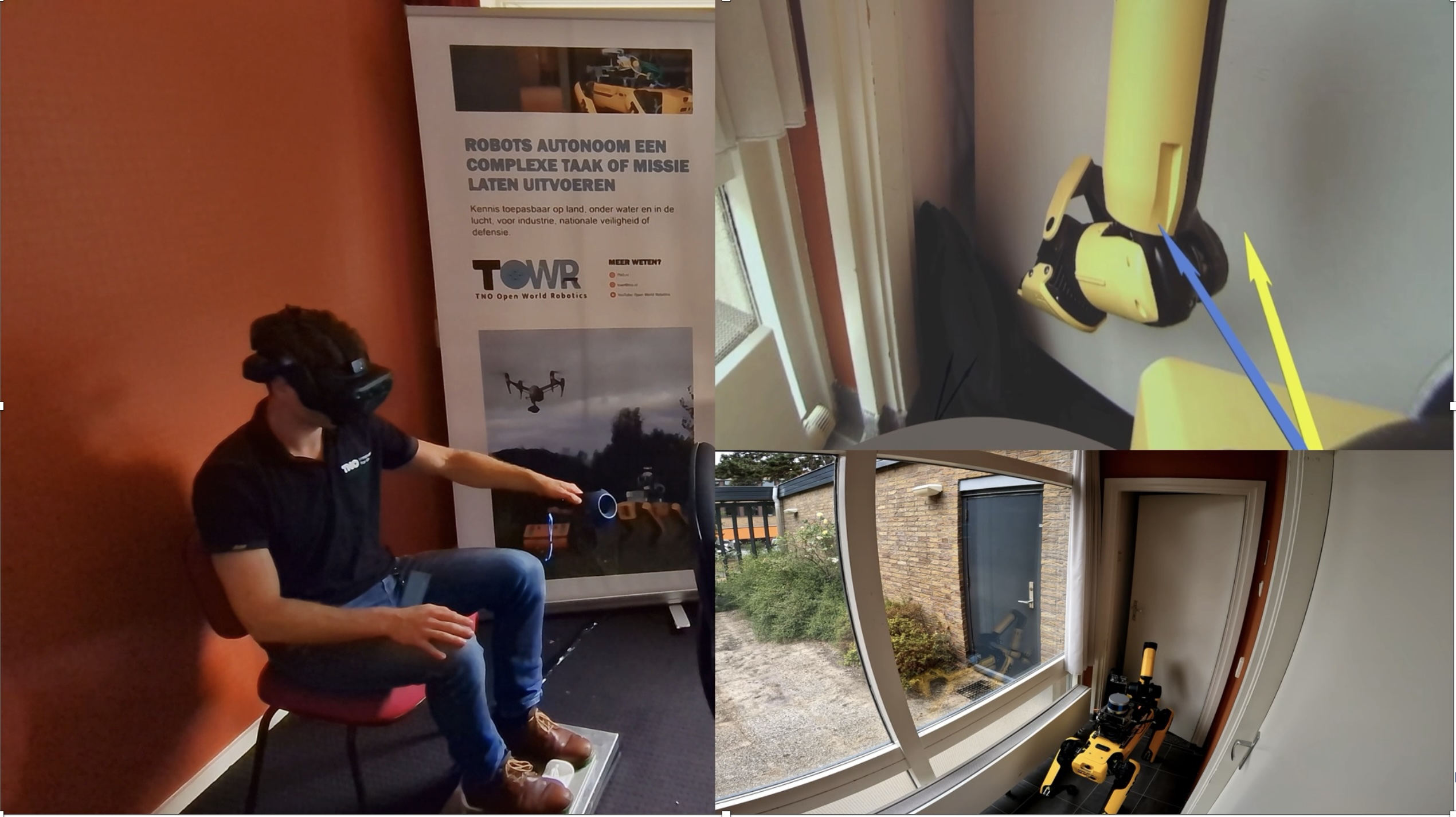}
    \caption{The setup for teleoperation. On the left the operator with VR goggles and VR controller, and his feet on the haptic plate. On the top right the view as the operator sees it. The bottom right shows the teleoperated robot.}
    \label{fig:teleop}
\end{figure}

\subsection{Immersive Teleoperation}
Whenever the operator feels the necessity to intervene, or when the perception server detects an object or situation where intervention is required, the system can switch to tele-operation mode. 
For this, the robot is equipped with a stereo pair of 180 degree fisheye lenses creating depth vision for the operator on the base of the arm. 
The two video streams are rendered through a Head Mounted Display (HTC Vive Pro Eye, HTC Corporation, Taiwan). 
The camera inputs are displayed through the Unity game engine on half domes matching the shape of the lens to compensate for distortion. 
The arm is controlled through end-effector position control. 
Input for the arm control is gathered with the VR controller (HTC Vive Pro Eye). 
Locomotion of the robot is controlled through a custom foot pedal, measuring the weight distribution (towards toes, or towards heels) and rotation of the operator feet.
\autoref{fig:teleop} gives an impression of the operator station and the first person VR view. 

%% file: src/experiments.tex
\section{Experimental Set-up}\label{sec:experiment}
In order to iterate with our stakeholders on the system requirements, an experiment was devised. This section describes the experimental robot platform and the setting of the experiment.

\subsection{Experimental Platform}\label{sec:experimental-platform}
\begin{figure}
    \centering
    \includegraphics[width=0.8\columnwidth]{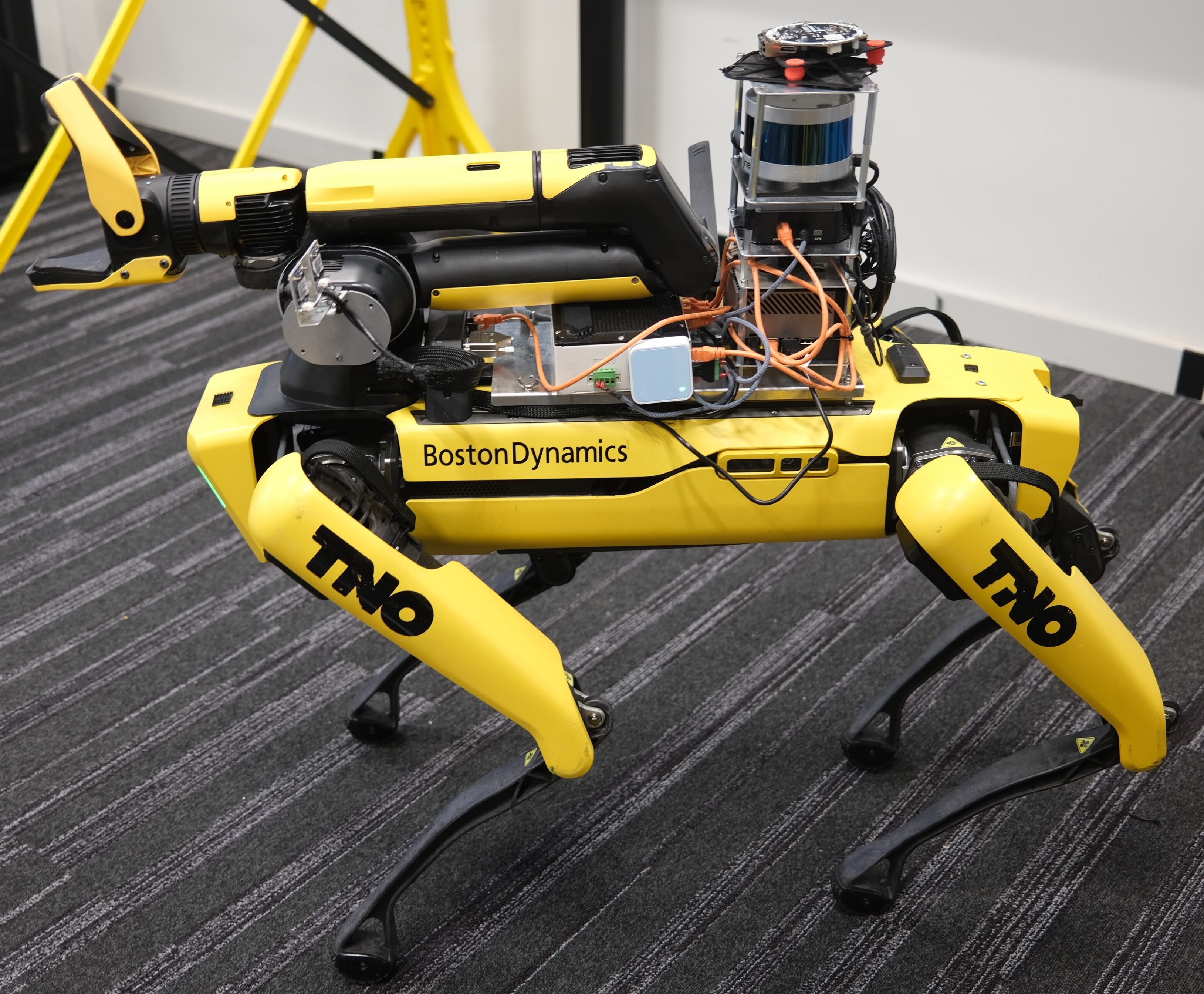}
     \caption{The experimental platform is a Spot robot with arm equipped with a custom backpack.}
    \label{fig:experimental-platform}
\end{figure}

The experimental robotic platform is a Spot robot with arm and a custom backpack, as shown in \autoref{fig:experimental-platform}. The backpack comprises a 4G-router (HMS), A LiDAR Velodyne VLP-16), IMU (microstrain), a microphone array (ReSpeaker) and a Nvidia Jetson PC (Jetson AGX Orin). At the base of the arm, two fisheye cameras (ELP) are mounted for use during teleoperation.
\autoref{fig:system-architecture} shows an overview of the software architecture, and on which hardware the two applications run.

\subsection{Experiment for Formative Evaluation of the Complete System}
An experiment was conducted at a mock-up drug lab normally used for training of the Dutch Police. 
The goal was twofold: testing the initial version of the system in a relevant environment and evaluating it with potential end users.

For the experiment, the participants from the Dutch Police were tasked with building situational awareness of the area and placing any detected dangerous objects into a safe container. Specifically, the mission consisted of the following tasks:
\begin{enumerate}
    \item Supervise the system while it autonomously searches for the 3 hazardous situations in the environment.
    \item Perform 2 immersive tele-operation interventions. A bottle representing a "toxic chemical" should be manipulated and placed securely in the waste box. 
\end{enumerate}

The total number of hazardous situations and required interventions were unknown to the police officers. 

\begin{figure}
    \begin{subfigure}{0.49\columnwidth}
      \centering
      \includegraphics[width=0.99\textwidth]{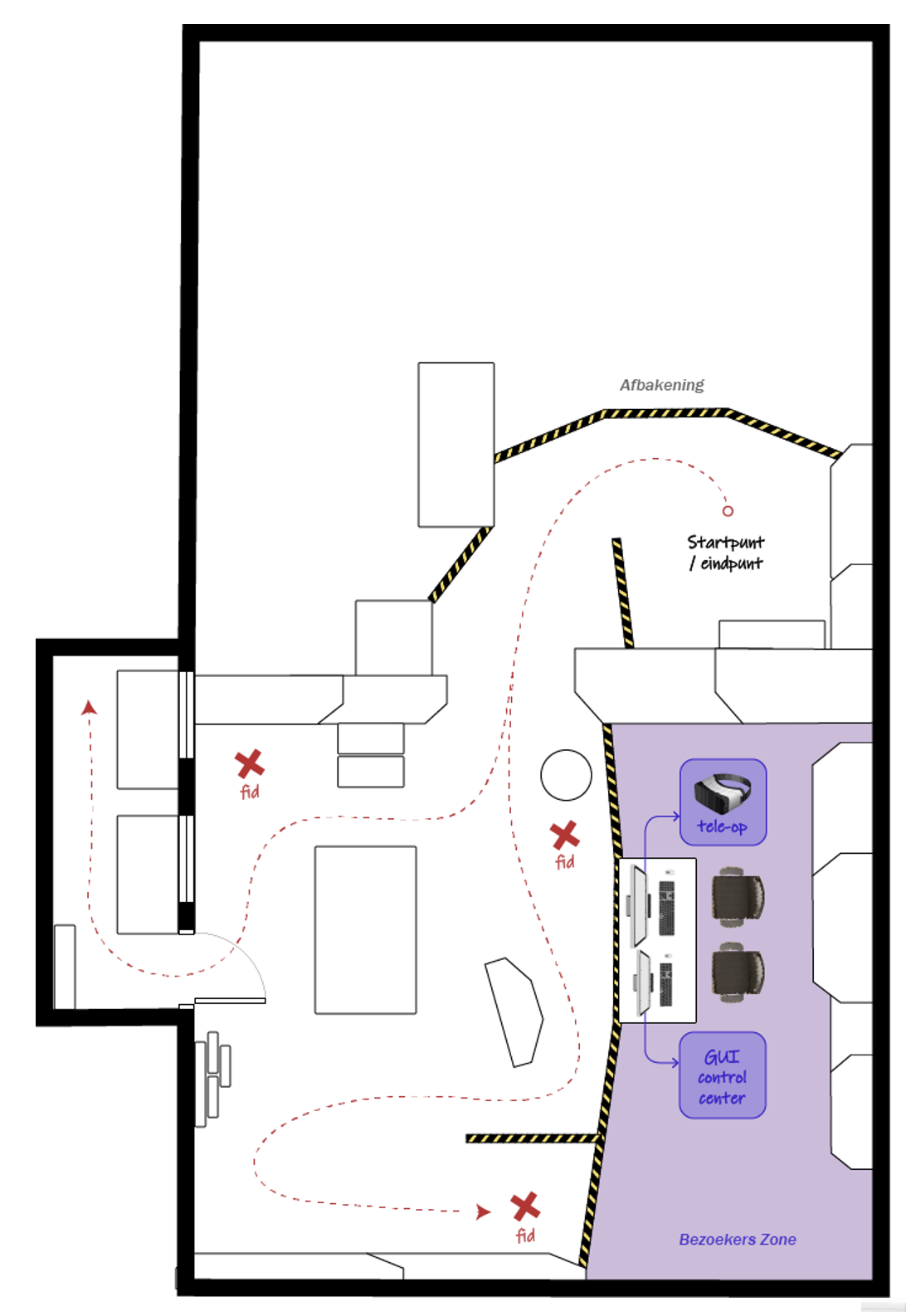}
    \end{subfigure}%
    \begin{subfigure}{0.49\columnwidth}
      \centering
      \includegraphics[width=0.99\textwidth]{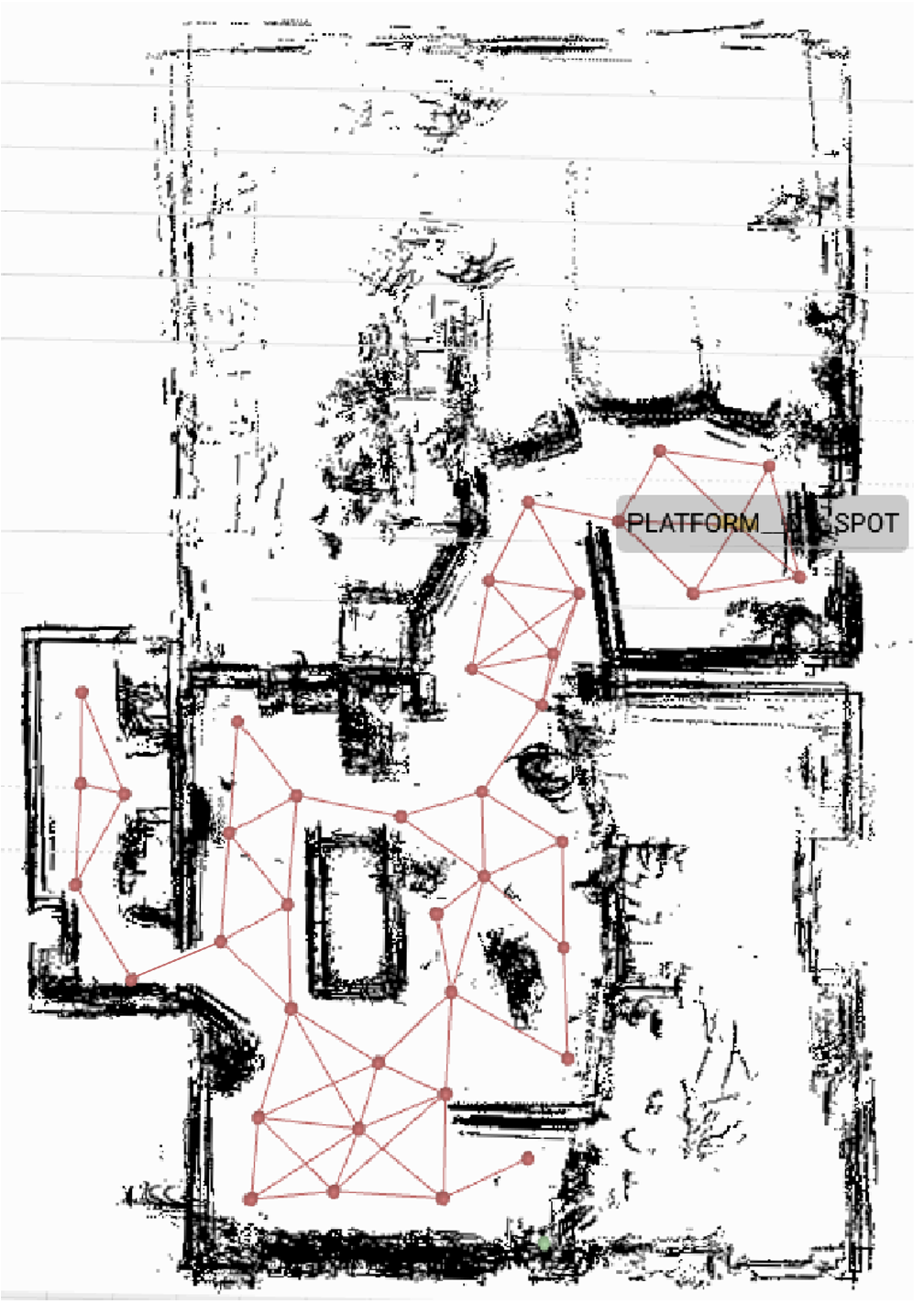}
    \end{subfigure}%
    \caption{Schematic drawing of the mock-up drug lab on the left. The resulting environment representation of the test location on the right.}
    \label{fig:drugs-lab}
\end{figure}

A map of the experimental environment used is shown on the left in \autoref{fig:drugs-lab}. 
It represents an unknown indoor cluttered environment, roughly 15x20 meters in size and included a separate room connected by a door already opened. The remote control station is envisioned to be at a distant location (e.g. police van or police control room), however, for practical reasons in this experiment it was placed nearby in the same room. 

The group of participants consisted of four officers from the Dutch National Police working in the discovery and dismantling of drug labs. The participants received a short 5 minute instruction on the controls of the tele-operation setup and an short explanation of the user interface before the start of the experiment.

%% file: src/results.tex
\section{Experimental Results}\label{sec:results}
This section will describe the results of the formative study we conducted together with the stakeholders. The experimental results contribute to an iteration of the system requirements.
We focus especially on how the Situational Graph should be adapted to optimally connect to the new requirements.

\subsection{Results Formative Evaluation with Stakeholders}

Feedback on the demonstrated system was overall positive. 
The envisioned task of building situational awareness was supported by the autonomous exploration and mapping and the task of placing dangerous objects in a safe container was supported by the identification of these objects and switching to tele-operation mode.
Operators appreciated the approach of adjustable levels of autonomy in itself, as it improved operator trust in tackling the dynamic environments in which they work by providing them with a means to act as a fallback for the robot, and seize control if needed. 

Even during a fully tele-operated mission with the system, the operators experienced benefits of the passive autonomy services running in the background:
\begin{itemize}
    \item The green frontier nodes helped in spotting missed and undiscovered rooms.
    \item When reaching the end of a room the operators could rely on the autonomous navigation to move the system to the start of an unexplored room or back to the start, rather than having to control the system there manually, this saved mental overhead.
    \item When experiencing communication issues they could rely on the system to autonomously return back into communication range.
\end{itemize}

They did note that the general approach of ``autonomy first, teleoperation second" was a big step compared to their current usage of robots, and felt more comfortable with a ``teleoperation first, autonomy second" approach.

%% file: src/conclusion.tex
\section{Conclusion}
Through the multidisciplinary process presented, we can iterate on robotic solutions for challenging unknown environments together with stakeholders in the national security domain.
We have shown that \sgraph{s} are powerful representations for storing actionable information. 
They are interpretable for humans while for robots they are directly usable for planning. 
By perceiving and storing actionable elements in the environment the used planning methods can simply be graph search, allowing the planning to occur at a high frequency while scaling to extensive areas of operation. 
Moreover, the representation allows the human operator to easily adjust the level of autonomy, and possibly switch to immersive teleoperation.

\subsection{Future Work}
The early and continuous evaluation of the system provides directions for future extensions of the Situational Graph.
Given the experimental results, a number of requirements were distilled for a subsequent development iteration. 

\begin{itemize}
    \item The robot should be able to pinpoint detected objects relevant to a specific mission in the graph.
    \item The robot should have specific execution modes for exploration: faster exploration, or slower but thorough.
    \item The robot framework should support deploying multiple robots that coordinate their efforts. 
    \item The robot framework should work with robots of differing type and make (interoperability). 
    \item The robot should be operated in an intuitive manner. 
\end{itemize}